# 2. Empowering sustainable finance with artificial intelligence: a framework for responsible implementation

*Georgios Pavlidis*

## I. Introduction

The global economy and societies around the world are facing a once-in-a-lifetime situation, as two major developments are underway and becoming collinear. On the one hand, financial markets are rapidly entering the era of environmental, social, and governance (ESG) investing. Market participants predict that investors' demand for more diverse instruments, such as green and ESG-linked loans, will increase in the coming years.[1] On the other hand, the artificial intelligence (AI) industry is experiencing close to exponential growth, and the impact of this new technology on business and society has already become visible. Indeed, the estimated value of the global AI market was $387.45 billion in 2022, and it is projected to reach $1,394.30 billion in 2029, with a compound annual growth rate of 20.1 per cent in this period.[2]

The aim of this chapter is to examine how these two processes – the rise of ESG investing (Section II) and the ascendance of AI technological innovation (Section III) – could be aligned and which risks and opportunities would emerge from this alignment. We argue that AI can help markets identify and price climate risks as well as set more ambitious ESG goals, yet there are serious risks in delegating sustainable finance decisions to AI. We further argue that developing new principles and rules for AI and ESG investing is necessary but prone to ambiguities and practical obstacles that could undermine norm-setting initiatives (Section IV). We conclude that implementing changes such as the use of AI in non-financial reporting requires a new sense of responsibility and the fine-tuning of the principles of legitimacy, oversight and verification, transparency, and explainability, along with international coordination in the context of AI and ESG investing (Section V).

---

[1] United Nations Environmental Program, 'State of Finance for Nature 2022' (UNEP Report, 1 December 2022) <https://www.unep.org/resources/report/state-finance-nature-2022> accessed 7 February 2024.

[2] Fortune Business Insights, 'AI Market Size to Reach USD 1394.30 Billion by 2029' (Market Research Report, April 2023) <https://www.fortunebusinessinsights.com/industry-reports/artificial-intelligence-market-100114> accessed 7 February 2024. According to other estimates, the market size of the AI industry was USD $93.5 billion in 2021 and is projected to expand at a compound annual growth rate of 38.1 per cent from 2022 to 2030; see Grand View Research, 'Artificial Intelligence Market Size, Share & Trends Analysis Report' (Market Analysis Report, April 2022) <https://www.grandviewresearch.com/industry-analysis/artificial-intelligence-ai-market> accessed 7 February 2024.

## II. Unlocking the potential of sustainable finance and navigating the pitfalls of responsible investing

The first trend worth examining is the increase in global awareness about sustainable investing – in particular, ESG, which extends the notion of sustainability beyond the need to protect the environment and mitigate climate change. The environmental component of ESG may include climate policies and policies governing energy use, waste, pollution, greenhouse gas (GHG) emissions, natural resource conservation, and so on.[3] The social component of ESG may include policies on diversity, inclusion, a community focus, social justice, and corporate ethics, such as ethical supply chains and prevention of questionable labour practices domestically and overseas.[4] The governance component may include the use of accurate and transparent accounting methods, corporate transparency, diversity in leadership, and accountability to shareholders.[5] In the ESG context, both policymakers and the markets shift from an exclusive focus on financial factors towards non-financial aspects when identifying risks and growth opportunities. Indeed, the importance of non-financial information in establishing a company's reputation and influencing the decisions made by investors is growing rapidly.[6]

ESG investing has become a clearly detectable and increasingly widespread trend in business, finance, and policymaking. According to Bloomberg, global ESG assets surpassed $35 trillion in 2020, and they may surpass $41 trillion by 2022 and $50 trillion

---

[3] Sachini Supunsala Senadheera et al., 'Scoring Environment Pillar in Environmental, Social, and Governance (ESG) Assessment' (2021) 7(1) Sustainable Environment 1.
[4] Elizabeth Pollman, 'Corporate Social Responsibility, ESG, and Compliance' in Benjamin van Rooij and D Daniel Sokol (eds), *The Cambridge Handbook of Compliance* (Cambridge University Press 2021) 662–672.
[5] Julie Linn Teigland, 'How can boards strengthen governance to accelerate their ESG journeys?' (EY Long-Term Value and Corporate Governance Survey, February 2022) <https://assets.ey.com/content/dam/ey-sites/ey-com/en_gl/topics/attractiveness/ey-long-term-value-and-corporate-governance-survey-february-2022.pdf> accessed 7 February 2024.
[6] Muhammad Naveed et al., 'Role of Financial and Non-Financial Information in Determining Individual investor Investment Decision: A Signaling Perspective' (2020) 9 South Asian Journal of Business Studies 261.

by 2025.[7] This will represent one-third of the assets under management globally. With regard to bond markets, there is already a sizeable $4 trillion ESG global debt market, which could swell to $15 trillion by 2025. ESG debt markets are dominated by European private and public issuers and investors, and this trend will be strengthened by the implementation of the colossal Green Deal of the European Union (EU).[8] With regard to ESG exchange-traded funds (ETF), cumulative assets in this category reached over $360 billion in 2021. This represents a small part (4 per cent) of global ETF assets, but the share of ESG in ETF annual flows is increasing rapidly, and investors expect that this expansion will continue in 2023 and beyond.[9] However, the trend in favour of ESG investment could be halted due to unforeseen events. For example, amid the Coronavirus Disease 2019 (COVID-19) pandemic, environmental protection and the fight against climate change were sidelined temporarily, as is often the case in times of crisis.[10] Nevertheless, as policymakers set in motion the recovery of national economies, green and sustainable development became a key element of economic recovery plans, especially at the level of the EU, with its Green Deal.[11]

The trend in favour of ESG investment is undoubtedly a positive one and paved with good intentions. However, the transition towards ever-stronger ESG performance contains some serious pitfalls. The three main ones are the lack of definitive, universal ESG metrics, the risk of so-called 'greenwashing', and the labour-intensiveness of ESG reporting.

---

[7]  Bloomberg, 'ESG May Surpass $41 Trillion Assets in 2022, But Not Without Challenges, Finds Bloomberg Intelligence' (Bloomberg Press Announcement, 24 January 2022) <https://www.bloomberg.com/company/press/esg-may-surpass-41-trillion-assets-in-2022-but-not-without-challenges-finds-bloomberg-intelligence/> accessed 7 February 2024.

[8]  In summer 2020, the European Council adopted the EU recovery plan and multiannual financial framework (MFF) for 2021–2027 with the objective of ensuring 'that the next MFF as a whole contributes to the implementation of the Paris Agreement'; Conclusions of the Special Meeting of the European Council (17–21 July 2020) 14. The EU intends to borrow €750 billion under NextGenerationEU, 30 per cent of which will be raised through green bonds.

[9]  BNP Paribas Asset Management, 'ESG ETF Barometer shows strong shift in investor expectations' (BNP Press Release, February 2023) <https://mediaroom-en.bnpparibas-am.com/news/bnp-paribas-asset-management-european-esg-etf-barometer-shows-strong-shift-in-investor-expectations-26f5-0fb7a.html> accessed 7 February 2024.

[10] Charlotte Burns, Peter Eckersley, and Paul Tobin, 'EU Environmental Policy in Times of Crisis' (2020) 27(1) Journal of European Public Policy 1; Yves Steinebach and Christoph Knill, 'Still an Entrepreneur? The Changing Role of the European Commission in EU Environmental Policymaking' (2017) 24(3) Journal of European Public Policy 429.

[11] Of course, there are significant risks and shortcomings. For example, the EU Green Deal does not focus on the social dimensions of sustainable development, and it fails to make sufficiently clear the cost-benefit effects for citizens. See Sanja Filipović, Noam Lior, and Mirjana Radovanović, 'The Green Deal – Just Transition and Sustainable Development Goals Nexus' (2022) 168 (13) Renewable and Sustainable Energy Reviews 112759.

First, ESG investing requires predetermined, commonly accepted, and uniformly applied ESG metrics. Currently, there is no definitive, universal set of ESG metrics, although there are various distinct frameworks and certifications that differ in scope, methodology, and acceptance by markets, for instance the Global Reporting Initiative (GRI), the Sustainability Accounting Standards Board (SASB) Standards, the Carbon Disclosure Project (CDP), and the International Organization for Standardization's ISO 14001 for environmental management systems. The variety of ESG business projects does not help, since different metrics will be relevant for numerous types of projects, from renewable energy and green infrastructure projects to payments for ecosystem services, biodiversity offsets, and pro-biodiversity and adaptation businesses. Depending on the project, ESG metrics may include carbon footprints, GHG emissions, year-over-year energy, waste variance, community impact, and pay gap, not all of which can be easily measured, and surely not with the same methodology. The volume of data and the complexity of the metrics are likely to increase, as will the variety of green projects that receive financing. Therefore, ensuring convergence will be a serious challenge for ESG metrics,[12] given the growing demand of investors for extra-financial data, including 'sustainable performance' and 'green performance' data, which are required to meet the fragmented international and regional standards.[13]

A second pitfall is the risk of 'greenwashing', that is to say, misleading information regarding how a product or service is environmentally sound.[14] A characteristic example of greenwashing was the 'Dieselgate' scandal in which the car giant Volkswagen admitted to cheating emissions tests by using proprietary software (the 'defeat device') and altering the performance of engines undergoing testing.[15] At the same time, the company was advertising the low-emissions features of its products through marketing campaigns.[16] In the specific context of ESG investing, greenwashing is the use of misinformation to gain investor confidence around a company's ESG claims.[17] Several cases of greenwashing have been reported in the financial industry, such as the recent advertising campaign of HSBC, which was banned by the United Kingdom (UK)'s advertising regulator for being 'misleading' about the company's contribution to the fight against climate change; the campaign in question outlined the bank's efforts to help its customers achieve 'net zero' emissions, but it omitted significant information about its role in financing fossil fuel firms and new oil and gas production.[18] In a troubling study conducted in 2021, the Economist examined the holdings of the world's 20 biggest ESG funds[19] and found that each of them held investments in fossil fuel producers, while others held stakes in oil producers, coal mining, gambling, alcohol, and tobacco.[20] The extensive range of ESG reporting enables corporations to prioritise one ESG metric over another, focussing on improving reported ESG metrics while ignoring unreported ones, thus masking or hiding the negative impacts of their actions.[21] To prevent greenwashing in ESG investing, reduce informational asymmetries, and ensure the accuracy and quality of extra-financial reporting, investors need to demand and markets need to develop reliable labels, certifications, and verifications to be carried out by independent third parties. The adoption of regulations imposing transparency and high standards in non-financial reporting can accelerate this development, as we will examine in Section IV of this study.

---

[12]  Luluk Widyawati, 'A Systematic Literature Review of Socially Responsible Investment and Environmental Social Governance Metrics' (2020) 29(2) Business Strategy and the Environment 619.

[13]  United Nations Environment Program, 'Driving Meaningful Data: Financial Materiality, Sustainability Performance and Sustainability Outcomes' (UNEP Finance/United Nations Global Compact, September 2020) 3.

[14]  Maria J Bachelet, Leonardo Becchetti, and Stefano Manfredonia, 'The Green Bonds Premium Puzzle: The Role of Issuer Characteristics and Third-Party Verification' (2019) 11(4) Sustainability 1098.

[15]  Edin Mujkic and Donald Klingner, 'Dieselgate: How Hubris and Bad Leadership Caused the Biggest Scandal in Automotive History' (2019) 21(4) Public Integrity 365.

[16]  Regulators in multiple countries reacted and began investigating Volkswagen. As a result, the stock price of the company fell in value, and key executives were suspended.

[17]  This term is rather broad, and there is a lack of shared understanding about its definition and taxonomy; see Soh Young In and Kim Schumacher, 'Carbonwashing: ESG Data Greenwashing in a Post-Paris World' in Thomas Heller and Alicia Seiger (eds), *Settling Climate Accounts* (Palgrave Macmillan 2021) 39–58.

A third challenge to ESG investing is the fact that extra-financial reporting is labour intensive.[22] Reporting on ESG projects requires collecting, analysing, and publishing information on ESG impacts, leading to complex and technical metrics. The costs of extra-financial reporting are recurring and, under current market practices, extend to the entire lifecycle of the project, as reporting takes place periodically, usually annually, prior to the full allocation of proceeds.[23] Moreover, several jurisdictions have enhanced reporting requirements, further increasing the costs. The increased workload and costs associated with ESG investing may deter investors, but new and promising methods and technologies, such as AI, have the potential to alleviate most of these constraints and even allow for the use of alternative data as an objective and reliable data source.

---

[18] Annabelle Liang, 'HSBC climate change adverts banned by UK watchdog' *BBC* (19 October 2022) <https://www.bbc.com/news/business-63309878> accessed 7 February 2024.

[19] The Economist, 'Sustainable finance is rife with greenwash. Time for more disclosure' (22 May 2021) <https://www.economist.com/leaders/2021/05/22/sustainable-finance-is-rife-with-greenwash-time-for-more-disclosure> accessed 7 February 2024.

[20] See also Tim Espiner, 'Big banks fund new oil and gas despite net zero pledges' *BBC* (14 February 2022) <https://www.bbc.com/news/business-60366054> accessed 7 February 2024. According to this article, big banks continue to fund oil and gas projects despite their net zero pledges and being part of the United Nations-led Net Zero Banking Alliance.

[21] Robert S Kaplan and Karthik Ramanna, 'How to Fix ESG Reporting' (2021) Harvard Business School Accounting & Management Unit Working Paper No. 22–005, <https://www.hbs.edu/ris/Publication%20Files/22-005revised_ed6ac430-c3ca-4ba6-b0be-ca48c549aaf2.pdf> accessed 7 February 2024. An interesting example given by these authors is a company that reports a decrease in GHG emissions from its truck fleet, which is an ESG metric that is measured and reported on; however, the same company does not report on another ESG metric, namely the use of indentured labour in mining minerals for electric vehicles' batteries, which may have worsened as a result of its actions.

[22] EU Technical Expert Group on Sustainable Finance, 'Report on EU green bond standard' (TEG Report Proposal for an EU Green Bond Standard, June 2019) 22.

The rise of ESG investing is associated with risks, like all new trends in the worlds of business and finance. Not surprisingly, there have been several proposals and norm-setting initiatives, but most are still under development, and they have still not matured into a comprehensive, coherent, and universally applied framework of rules and principles. Currently, principles and rules for ESG investing can be found at three levels: a) global goals, such as the United Nations (UN) Sustainable Development Goals, Principles for Responsible Investment, Greenhouse Gas Protocol, and Paris Climate Agreement; b) voluntary reporting frameworks, as in the context of the GRI, the ISO, the Science Based Targets Initiative (SBTI), the Climate Disclosure Standards Board (CDSB), the SASB, and the CDP; c) mandatory regulations at the national and EU levels, such as the EU Corporate Sustainability Reporting Directive (CSRD), Sustainable Finance Disclosure Regulation (SFDR), and EU Taxonomy.[24] The challenge for policymakers and standard-setting bodies will be to promote the global consolidation of ESG reporting governance by overcoming the differences between their respective reporting philosophies.[25] Until this happens, the challenge for issuers and investors will be navigating the complex landscape and finding the sweet spots between ESG reporting initiatives that are relevant to their projects.

---

[23] Deloitte, 'Thinking Allowed: The future of corporate reporting' (Deloitte, July 2016) 8 <https://www2.deloitte.com/content/dam/Deloitte/ch/Documents/audit/ch-en-audit-thinking-allowed-future-corporate-reporting.pdf> accessed 7 February 2024.

[24] Directive (EU) 2022/2464 of the European Parliament and of the Council of 14 December 2022 amending Regulation (EU) No 537/2014, Directive 2004/109/EC, Directive 2006/43/EC and Directive 2013/34/EU, as regards corporate sustainability reporting [2022] OJ L 322/15 of 16.12.2022; Regulation (EU) 2019/2088 of the European Parliament and of the Council of 27 November 2019 on sustainability-related disclosures in the financial services sector [2019] OJ L 317/1 of 09.12.2019; Regulation (EU) 2020/852 of the European Parliament and of the Council of 18 June 2020 on the establishment of a framework to facilitate sustainable investment, and amending Regulation (EU) 2019/2088 [2020] OJ L 198/13 of 22.06.2020.

[25] Adam Sulkowski and Ruth Jebe, 'Evolving ESG Reporting Governance, Regime Theory, and Proactive Law: Predictions and Strategies' (2022) 59(3) American Business Law Journal 449.

## III.   The rise of artificial intelligence, its impact on finance, and the risks of hype in the absence of regulation

According to the Organisation for Economic Co-operation and Development (OECD), 'an AI system is a machine-based system that can, for a given set of human-defined objectives, make predictions, recommendations, or decisions influencing real or virtual environments … with varying levels of autonomy'.[26] Almost identically, the EU defines an AI system as 'software that … can, for a given set of human-defined objectives, generate outputs such as content, predictions, recommendations, or decisions influencing the environments they interact with'.[27] AI refers to the ability of a computer to accomplish tasks that require human intelligence and discernment and, for this reason, are usually executed by humans. AI encompasses subfields, such as machine learning and deep learning, in which the algorithms use input data to learn, draw inferences, and adapt without explicit instructions.[28] Numerous AI applications have already been developed and many others are underway, from manufacturing robots, self-driving cars, smart assistants, and marketing chatbots to AI applications in healthcare, agriculture, and finance.[29]

These exciting prospects have attracted the fervent interest of investors worldwide, and as a result, global spending on AI systems reached $118 billion in 2022 and is expected to surpass $300 billion in 2026, which will represent a massive compound annual growth rate of 26.5% over this period.[30] The banking and retail industries are expected to make the most significant investments in AI, collectively comprising around 25 per cent of all AI expenditure globally.[31] It is also projected that by 2030, AI will have the potential to add approximately $15.7 trillion to the global economy (a 14

---

[26]   OECD, 'Recommendation of the Council on Artificial Intelligence' (OECD Recommendation of the Council on Artificial Intelligence OECD/LEGAL/0449, 2019); See also IBM, 'What is Artificial Intelligence?' (3 June 2020) <https://www.ibm.com/topics/artificial-intelligence#anchor-1174800854> accessed 7 February 2024, according to which 'artificial intelligence is a field that combines computer science and robust datasets to enable problem-solving'.

[27]   Article 3 par. 1, Proposal for a Regulation laying down harmonised rules on artificial intelligence (Artificial Intelligence Act), COM(2021) 206 final.

[28]   Several classifications have been proposed: AI can be reactive, have limited memory, have a theory of mind, and be self-aware. Authors also distinguish between artificial narrow intelligence, artificial general intelligence, and artificial super intelligence. Among numerous interesting works, see Stuart Russell and Peter Norvig, *Artificial Intelligence: A Modern Approach* (4th edn, Pearson Education 2021).

[29]   Nicole Radziwill, *Connected, Intelligent, Automated: The Definitive Guide to Digital Transformation and Quality 4.0* (Quality Press 2020) 49.

[30]   International Data Corporation, 'Worldwide Artificial Intelligence Spending Guide' (IDC Spending Guide, September 2022). According to the IDC, 27 per cent of the growth in AI by 2023 will be in hardware, which includes devices such as AI-powered smart speakers; 38 per cent of the growth in AI will be in services, such as information technology; and 35 per cent will be in software.

[31]   Ibid.

per cent increase in global GDP), which is more than the combined output of China and India at present. Out of this total amount, an estimated $6.6 trillion is expected to result from a boost in productivity, while approximately $9.1 trillion is expected to result from the consumption side.[32]

In the field of finance, AI can be used both for portfolio management and client enablement, as well as for improving front-, middle-, and back-office efficiency. For example, AI applications can help evaluate management sentiment by reading earnings transcripts and generating automated insights. AI can also help identify non-intuitive relationships between assets, market indicators, and alternative datasets, such as social media, weather forecasts, and container ship movements. Furthermore, AI can automate functions in financial operations intelligence, monitor suspicious transactions, generate reports, and respond to employee or investor queries using machine learning and chatbots.[33]

Moreover, applications can combine all elements of the so-called 'BIA Trinity', that is, blockchain, the Internet of Things (IoT), and AI, to revolutionise numerous aspects of our lives, including the world of sustainable finance.[34] This will allow data from the real economy (e.g., data from sensors) in a sustainable project to be uploaded to the blockchain directly and in real time.[35] Impact reports could then be compiled with the use of AI to analyse complex data to be ultimately delivered to the digital wallet of the investor, an independent certification agency, or the oversight authorities. This will constitute a shift from manual to automated reporting by harvesting recognised metrics from assets, codifying them as data tokens, and communicating them to multiple stakeholders. Automatic reporting will increase the reliability and traceability of projects' performance. Moreover, it has been estimated that such digitalisation could reduce the average cost of data gathering (including the cost of IoT devices), data aggregation, and reporting for the full lifecycle of a sustainable project by up to ten times.[36]

---

[32] PwC, 'Sizing the prize: What's the real value of AI for your business and how can you capitalise?' (PWC Global Artificial Intelligence Study, 2017) <https://www.pwc.com/gx/en/issues/analytics/assets/pwc-ai-analysis-sizing-the-prize-report.pdf> accessed 7 February 2024.

[33] Deloitte, 'Artificial Intelligence: The next frontier for investment management firms' (Deloitte, 2022) <https://www.deloitte.com/global/en/Industries/financial-services/perspectives/ai-next-frontier-in-investment-management.html> accessed 7 February 2024.

[34] Darius Nassiry, 'The Role of Fintech in Unlocking Green Finance: Policy Insights for Developing Countries' (2018) Asian Development Bank Institute, ADBI Working Paper Series No. 88310 <https://www.adb.org/sites/default/files/publication/464821/adbi-wp883.pdf> accessed 7 February 2024.

[35] Wanli Chen and Qianxia Wang, 'The Role of Blockchain for the European Bond Market' (2020) Frankfurt School Blockchain Center, FSBC Working Paper, 11 <http://explore-ip.com/2020_The-Role-of-Blockchain-for-the-European-Bond-Market.pdf> accessed 7 February 2024.

It is a positive sign that many economic sectors, such as transport, energy, and water, already employ IoT devices and harvest data automatically, which can be useful for extra-financial reporting.[37] Nevertheless, the exact metrics for automated reporting must be determined – that is, which specific data are to be collected, analysed, and reported. Once this has been determined, it will be possible to generate impact indexes for one or more projects by combining relevant indicators, such as measurements of air quality or $CO_2$, from multiple monitoring points over time. To deal with the lack of well-structured data formats and the huge volume of datasets gathered automatically, AI tools and advanced analytics can be employed to organise and evaluate data volumes. For example, an application developed by Global Mangrove Trust uses AI to monitor forest density, a particularly complex set of data, and combines it with TreeCoin rewards, which are exchangeable for goods.[38] AI can be deployed in similarly large and complex datasets, from the evapotranspiration necessary for the preservation of wetlands ecosystems[39] to geoinformatics in smart, sustainable agriculture.[40] The use of AI and digitalisation in sustainable finance has the potential to economise time and costs, facilitate the scaling-up of sustainable investments, improve the quality of reporting, and facilitate aggregating reporting, allowing for follow-up innovations, such as the fragmentation of green asset ownership and the aggregation of smaller assets and projects into a bond.

---

[36] HSBC Sustainable Digital Finance Alliance, 'Blockchain: Gateway for Sustainability Linked Bonds' (HSBC Centre of Sustainable Finance, September 2019) 17 <https://www.sustainablefinance.hsbc.com/-/media/gbm/reports/sustainable-financing/blockchain-gateway-for-sustainability-linked-bonds.pdf> accessed 7 February 2024.

[37] Sandro Nizetic et al., 'Internet of Things (IoT): Opportunities, Issues and Challenges towards a Smart and Sustainable Future' (2020) 274 Journal of Cleaner Production 274; In Lee and Kyoochun Lee, 'The Internet of Things (IoT): Applications, Investments, and Challenges for Enterprises' (2015) 58(4) Business Horizons 431–40; HSBC Sustainable Digital Finance Alliance, 'Blockchain: Gateway for Sustainability Linked Bonds' (HSBC Centre of Sustainable Finance, September 2019) 20 <https://www.sustainablefinance.hsbc.com/-/media/gbm/reports/sustainable-financing/blockchain-gateway-for-sustainability-linked-bonds.pdf> accessed 7 February 2024.

[38] TreeCoin, *Tree Coin White Paper* (White Paper, November 2020) 23.

[39] Francesco Granata, Rudy Gargano, and Giovanni de Marinis, 'Artificial Intelligence Based Approaches to Evaluate Actual Evapotranspiration in Wetlands' (2020) 703 Science of The Total Environment 135653.

[40] Abhishek Singh et al., 'Geoinformatics, Artificial Intelligence, Sensor Technology, Big Data: Emerging Modern Tools for Sustainable Agriculture' in Pavan Kumar et al. (eds), *Sustainable Agriculture Systems and Technologies* (John Wiley & Sons Ltd 2022) 295–313.

Nevertheless, faced with the near-exponential pace of progress in AI, including AI used for sustainability, serious concerns have been voiced regarding potential risks[41] and the need for more regulation and oversight, which we will examine in the following section.

## IV.   Developing new principles and rules for AI: a journey in unchartered territory

In 2014, the billionaire investor Elon Musk told an audience at the Massachusetts Institute of Technology (MIT) the following: 'I'm increasingly inclined to think that there should be some regulatory oversight, maybe at the national and international level, just to make sure that we don't do something very foolish. I mean with artificial intelligence, we're summoning the demon.'[42] The deployment of AI raises legitimate concerns about cyber resilience; data security, quality, consistency, completeness, and privacy; the risk of bias; and several other risks.[43] For this reason, new sets of principles and rules have been proposed to mitigate risks and guide responsible digital innovation in AI.

Among several other interesting initiatives introduced by governments, industry, and other stakeholders,[44] we can look at the OECD's 'Recommendation' on AI.[45] This non-binding instrument treats AI as a general-purpose technology – that is, not in the specific context of ESG investing – and puts forward five key principles,[46] which are worth discussing. Principle 1 (inclusive and sustainable growth) highlights the potential for trustworthy AI to contribute to overall growth and prosperity for all – individuals, society, and the planet – and advance global development objectives. In the ESG context, this means ensuring that the development and use of AI tools align with sustainability goals, social inclusivity, and environmental responsibility. Therefore, Principle 1 aims to enhance the contribution of AI to long-term value creation, sustainable development, and positive societal impacts. Principle 2 (fairness and rule of law) states that AI systems should be designed in a way that respects the rule of law, human rights, democratic values, and diversity, and should include appropriate safeguards to ensure a fair and just society. The main difficulty in this situation is creating efficient systems for informed oversight. Principle 3 (transparency and explainability) aims to ensure transparency and responsible disclosure around AI systems so that people understand when they are engaging with them and can challenge outcomes. In the context of ESG, this requires that the methodology and outcomes of an AI model be properly explained and communicated to the oversight authorities and/or independent certification agencies. Principle 4 (safety) provides that AI systems must function in a robust, secure, and safe way throughout their lifetimes, and potential risks should be continually assessed and managed. This concept closely resembles the EU's more intricate, risk-based approach, as outlined in the EU AI Act. In the ESG context, understanding and mitigating risks associated with AI technologies is crucial for ensuring alignment with sustainability goals and avoiding negative impacts on stakeholders and the environment. To illustrate, consider the hypothetical situation where an investment firm utilises an AI-powered trading algorithm designed to prioritise investments in companies with high ESG scores. The algorithm is programmed to buy or sell stocks based on real-time ESG data and market trends. However, due to a technical glitch or unforeseen data irregularity, the algorithm mistakenly interprets negative news about a socially responsible company as positive, leading it to heavily invest in that company's stocks. Such scenarios and risks must be mitigated promptly and effectively. Finally, Principle 5 (accountability) proposes that organisations and individuals developing, deploying, or operating AI systems should be held accountable for their proper functioning. In the ESG context, any organisation that undertakes ESG reporting remains responsible for ensuring compliance with the applicable legislative provisions and reporting standards, and it cannot invoke the use of an AI system to shield itself against liability.

---

[41]   Rohit Nishant, Mike Kennedy, and Jacqueline Corbett, 'Artificial Intelligence for Sustainability: Challenges, Opportunities, and a Research Agenda' (2020) 53 International Journal of Information Management 102104.

[42]   Gregory Wallace, 'Elon Musk warns against unleashing artificial intelligence "demon"' *CNN Business* (26 October 2014) <https://money.cnn.com/2014/10/26/technology/elon-musk-artificial-intelligence-demon/> accessed 7 February 2024.

[43]   Matthew U Scherer, 'Regulating Artificial Intelligence Systems: Risks, Challenges, Competencies, and Strategies' (2016) 29(2) Harvard Journal of Law & Technology 353.

[44]   Jessica Morley et al., 'From What to How: An Initial Review of Publicly Available AI Ethics Tools, Methods and Research to Translate Principles into Practices' (2020) 26 Science and Engineering Ethics 2141.

[45]   OECD, 'Recommendation of the Council on Artificial Intelligence' (OECD Recommendation of the Council on Artificial Intelligence OECD/LEGAL/0449, 2019).

[46]   Roger Clarke, 'Principles and Business Processes for Responsible AI' (2019) 35(4) Computer Law & Security Review 410.

These principles have influenced national and regional norm-setting initiatives, the most prominent and comprehensive of which is the AI Act of the EU. This legislation, adopted in December 2023, has yet to be published in the Official Journal of the European Union as of February 2024. This is a 'horizontal' rather than sector-specific piece of legislation, and it may overlap with several legislative provisions, such as those related to cybersecurity and data protection.[47] As a result, inconsistencies may emerge that negatively affect legislative quality and regulatory certainty. The EU Act follows a risk-based approach to AI, where legal intervention is tailored to levels of risk.[48] To that end, the AI Act distinguishes between AI systems posing (i) unacceptable risk,

---

[47] Arthur Bogucki et al., 'The AI Act and emerging EU digital acquis: Overlaps, gaps and inconsistencies' (CEPS In-Depth Analysis, September 2022) <https://cdn.ceps.eu/wp-content/uploads/2022/09/CEPS-In-depth-analysis-2022-02_The-AI-Act-and-emerging-EU-digital-acquis.pdf> accessed 7 February 2024.

[48] Luciano Floridi, 'The European Legislation on AI: A Brief Analysis of Its Philosophical Approach' in Jakob Mökander and Marta Ziosi (eds), *The 2021 Yearbook of the Digital Ethics Lab* (Springer 2022) 1–8.

(ii) high risk, (iii) limited risk, and (iv) low or minimal risk.[49] Under this approach, AI applications would be regulated only as strictly necessary to address specific levels of risk. Like the OECD Principles, the EU AI Act aims to ensure the transparency and explainability of AI. Datasets must be free of errors, and humans must be able to 'fully understand' how AI systems work. Nevertheless, sometimes even AI creators do not fully understand how AI programs arrive at their conclusions, while tech companies are uncomfortable about giving external auditors or regulators access to their source code and algorithms. Despite some loopholes and exceptions (e.g., facial recognition by the police is banned unless the images are captured with a delay or the technology is being used to find missing children), the EU AI Act could be used as a template for other jurisdictions,[50] as was the EU General Data Protection Regulation (GDPR) – the block's most famous tech export – which has been copied everywhere from California to India.

With the adoption of the EU AI Act, a significant transition has occurred from relying on soft-law principles to implementing hard-law regulations. This shift is deemed timely and fitting, especially in light of the numerous risks associated with the rapid advancement of AI technologies. As society increasingly integrates AI into various sectors, including sustainable finance, there is a necessity for robust oversight. Therefore, the next logical progression involves the creation of a comprehensive supranational authority dedicated to overseeing AI. This authority would be vested with powers and responsibilities, encompassing policy development, threat assessment, risk management, standard-setting, monitoring, and enforcement. Such an entity is essential to ensure that AI is deployed in a manner that prioritises ethical considerations and mitigates potential harms in many areas, including sustainable finance. The EU AI office or European AI Board, as part of their broader mandate, will play a key role in overseeing AI applications within this realm too, thus preventing potential negative impacts such as biased decision-making, unsustainable investments, and environmental degradation. Furthermore, the supranational AI authority will promote the establishment of clear guidelines and common standards for AI applications, including in sustainable finance, fostering trust and confidence among stakeholders, investors, and consumers.

Of course, not all jurisdictions will follow the model of the EU AI Act.[51] In the United States of America (USA), the National Artificial Intelligence Initiative Act of 2020 follows a non-interfering approach to AI with the goal of fostering research and design in this area.[52] One should not expect the US Congress to act anytime soon on AI regulation, since there is little chance such action would garner support from US Senators or House Representatives, although this risks creating 'conditions for a race to the bottom in irresponsible AI' in the US.[53] Indeed, a 2022 report by the US Federal Trade Commission raised serious concerns that even well-intended AI uses can be inaccurate, biased, and discriminatory, enabling increasingly invasive forms of surveillance.[54] In October 2023, President Biden issued an Executive Order on AI outlining seven principles for the development and use of AI across the federal government.[55] These principles emphasise fairness, non-discrimination, accountability, and public participation, but don't directly impose regulations, and it is not clear how and when federal agencies will extend their authority into the AI space under the Executive Order. In the UK, the 2021 National AI Strategy and the 2022 AI Regulation Policy Paper aimed to promote innovation and establish a risk-based approach to AI.[56] The UK does not have specific laws designed to regulate AI, but it is partially regulated through a combination of legal and regulatory requirements created for other purposes that also apply to AI technologies. Based on these requirements, some regulators in the UK, such as the Information Commissioner's Office and the Equality and Human Rights Commission, have taken steps to promote responsible AI use by making AI a strategic priority in their plans and issuing guidance on how existing legislation governs the use of new technologies, such as AI, in automated decision-making.[57]

---

[49] Under Art. 5, unacceptable risk, such as the use of AI for social scoring, is prohibited. Under Art. 6 & ss, high risk, such as the use of AI in education, justice, or immigration, requires a conformity assessment. Under Art. 52, limited risk, such as the use of chatbots and deepfakes, requires respect for transparency. Under Art. 69, low or minimal risk, such as the use of AI in video games and spam filters, requires the adoption and implementation of a code of conduct.

[50] Anu Bradford, *The Brussels Effect: How the European Union Rules the World* (Oxford University Press New York 2020).

[51] Corinne Cath et al., 'Artificial Intelligence and the "Good society": The US, EU, and UK Approach' (2018) 24(2) Science and Engineering Ethics 505.

---

[52] There is also an EU–US tech partnership (the Trade and Technology Council) designed for mutual understanding of the principles underlying trustworthy and responsible AI; see <https://commission.europa.eu/strategy-and-policy/priorities-2019-2024/stronger-europe-world/eu-us-trade-and-technology-council_en> accessed 7 February 2024.

[53] Andrew Ross Sorkin et al., 'Why lawmakers aren't rushing to police AI' *New York Times* (3 March 2023) <https://www.nytimes.com/2023/03/03/business/dealbook/lawmakers-ai-regulations.html> accessed 7 February 2024.

[54] Federal Trade Commission, 'Combatting Online Harms Through Innovation' (Federal Trade Commission Report to Congress, September 2022) <https://www.ftc.gov/reports/combatting-online-harms-through-innovation> accessed 7 February 2024.

[55] Joseph R Biden Jr, 'Executive Order on the Safe, Secure, and Trustworthy Development and Use of Artificial Intelligence' (The White House, 30 October 2023) <https://www.whitehouse.gov/briefing-room/presidential-actions/2023/10/30/executive-order-on-the-safe-secure-and-trustworthy-development-and-use-of-artificial-intelligence/> accessed 7 February 2024.

[56] Department for Digital, Culture, Media & Sport and Damian Collins MP, 'UK sets out proposals for new AI rulebook to unleash innovation and boost public trust in the technology' (UK Government Press Release, 18 July 2022) <https://www.gov.uk/government/news/uk-sets-out-proposals-for-new-ai-rulebook-to-unleash-innovation-and-boost-public-trust-in-the-technology> accessed 7 February 2024. See also Data Protection and Digital Information Bill 143 2022–23 (HL Bill 30, as introduced).

[57] For example, UK data protection law includes requirements for 'automated decision-making' and the processing of personal data that cover the development and training

## V.   Putting it all together: implementing changes and a new sense of responsibility

AI can foster the growth of ESG investing in multiple ways. It can help analyse the performance of sustainable investments in different asset classes using novel and alternative datasets. It can be combined with new technologies, including distributed ledgers, the IoT, and smart contracts, to ensure greater data quality, consistency, and comparability. AI can also measure and track ESG-related risks and impacts facing companies and investor portfolios. Nevertheless, serious risks remain, such as the risk of greenwashing, challenges to cybersecurity and privacy, the lack of coordination between regulators, and the use of adversarial AI. For these reasons, we argue in favour of a new sense of responsibility and for respecting some core principles governing the use of AI in the context of ESG investing. By merging established ESG principles and standards with emerging standards and principles specific to AI, such as those outlined by the OECD and EU frameworks, we arrive at the following core principles.

First, legal provisions and rigorous standards should apply both to AI processes and ESG decision-making. Second, external oversight and verification mechanisms need to ensure respect for regulations and standards, both for AI and ESG. Third, ESG decisions made by AI tools should be intelligible and explicable to investors, regulators, and oversight mechanisms in accordance with the principles of transparency and explainability. Fourth, regulation, oversight, and verification must function both at the national and international levels to cover cross-border cases. Indeed, regulating technological and financial innovation and imposing restrictions at the national level alone would be ineffective, since companies would simply opt for investing, doing research, and developing their products and services in jurisdictions that lack such restrictions.[58] In any event, regulatory intervention requires consultation with and the involvement of all major stakeholders (big tech companies, start-ups, financial markets, as well as academic institutions, non-governmental organisations, and civil

---

of AI; the Online Safety Bill (HL Bill 87, 2022-23(Rev)) also contains provisions relating to the design and use of algorithms. See the UK Secretary of State for Digital, Culture, Media and Sport, 'Establishing a pro-innovation approach to regulating AI' (UK Government Policy Paper, 18 July 2022) <https://www.gov.uk/government/publications/establishing-a-pro-innovation-approach-to-regulating-ai/establishing-a-pro-innovation-approach-to-regulating-ai-policy-statement> accessed 7 February 2024.

[58]   In an interview with the Washington Post in 2015, Dr Peter H Diamandis, pioneer in the field of innovation and incentive competitions, correctly pointed out that 'the notion that the government can regulate against [technological innovation] is a fallacy. If the government regulates against use of drones or stem cells or artificial intelligence, all that means is that the work and the research leave the borders of that country and go someplace else'; see Jena McGregor, 'How to go after big, bold goals' *Washington Post* (19 February 2015) <https://www.washingtonpost.com/news/on-leadership/wp/2015/02/19/the-xprize-founders-how-to-guide-for-going-after-big-bold-goals/> accessed 7 February 2024.

society) in the development of new principles and rules. By working collaboratively to establish clear principles and standards for AI in the context of ESG investing, we can ensure that these technologies are used to create positive social and environmental outcomes, mitigate potential risks, and foster a new era of responsible and sustainable finance powered by AI.